\documentclass[10pt,letterpaper]{article}

\usepackage{cogsci_template}
\usepackage{cogsci}
\cogscifinalcopy
\hypersetup{
    citecolor=purple
}
\usepackage{CJKutf8}
\usepackage{multirow}
\usepackage{array}

\acrodef{llm}[LLM]{Large Language Model}
\acrodef{lm}[LM]{Language Model}
\acrodef{ai}[AI]{Artificial Intelligence}
\acrodef{weat}[WEAT]{Word-Embedding Association Test}

\newcommand{\han}{Han Chinese}
\newcommand{\minor}{Ethnic Minorities}
\newcommand{\woman}{Woman}
\newcommand{\man}{Man}

\title{Word Embeddings Track Social Group Changes Across 70 Years in China}

\author{
    \begin{tabular}{c c c}
        \bf Yuxi Ma$^{1,\,\textrm{\Letter}}$\quad{}\quad{} & \bf Yongqian Peng$^{1,2}$\quad{}\quad{} & \bf Yixin Zhu$^{1,\,\textrm{\Letter}}$\\
        \normalfont yxma@stu.pku.edu.cn\quad{}\quad{} & \normalfont yqpeng@stu.pku.edu.cn\quad{}\quad{} & \normalfont yixin.zhu@pku.edu.cn
    \end{tabular}\vspace{6pt}
    \\%
    \begin{tabular}{l}
        \small $^1$ Institute for Artificial Intelligence, Peking University \quad{} $^2$ Yuanpei College, Peking University \quad{} $\textrm{\Letter}$ corresponding authors
    \end{tabular}
}

\begin{document}
\begin{CJK*}{UTF8}{gbsn}
\maketitle

\begin{abstract}
Language encodes societal beliefs about social groups through word patterns. While computational methods like word embeddings enable quantitative analysis of these patterns, studies have primarily examined gradual shifts in Western contexts. We present the first large-scale computational analysis of Chinese state-controlled media (1950-2019) to examine how revolutionary social transformations are reflected in official linguistic representations of social groups. Using diachronic word embeddings at multiple temporal resolutions, we find that Chinese representations differ significantly from Western counterparts, particularly regarding economic status, ethnicity, and gender. These representations show distinct evolutionary dynamics: while stereotypes of ethnicity, age, and body type remain remarkably stable across political upheavals, representations of gender and economic classes undergo dramatic shifts tracking historical transformations. This work advances our understanding of how officially sanctioned discourse encodes social structure through language while highlighting the importance of non-Western perspectives in computational social science.

\end{abstract}

\section{Introduction}

Language serves as both a mirror and an architect of social reality. Through evolving linguistic representations, we can trace the shifting contours of power dynamics, cultural transformations, and social hierarchies that define human societies \citep{bourdieu1991language,fairclough2013language,cameron2012verbal}. These patterns do more than passively reflect change---they actively shape how societies perceive, value, and position different social groups within their hierarchical structures.

The computational revolution in linguistics, particularly the advent of word embedding representations, has transformed our capacity to analyze these linguistic patterns at unprecedented scales \citep{caliskan2017semantics,michel2011quantitative}. Recent studies leveraging word embeddings have revealed deep insights into systematic biases and evolving social attitudes within English-language corpora \citep{caliskan2017semantics,charlesworth2022historical,garg2018word}. Notably, a study of 72 social groups across a century of American English \citep{charlesworth2023identifying} found that while descriptive words changed, the underlying structure of group representations remained stable. However, this expanding field faces two significant limitations. First, our understanding remains largely confined to Western contexts, with systematic analyses of diachronic semantic evolution in other cultural spheres notably absent. Second, while we have extensive documentation of gradual social changes in relatively stable societies, we lack insight into how linguistic representations evolve during periods of radical social transformation---moments when traditional power structures face reorganization.

The People's Republic of China, since 1949, presents an unparalleled natural experiment for studying such linguistic evolution during rapid social change. The period 1950-2019 encompassed major shifts: the Cultural Revolution's (1966-1976) restructuring of social hierarchies, market reforms initiated by the 1978 Reform and Opening Up policy, and extensive rural-to-urban migration \citep{xie2011evidence}. These transformations did not merely modify existing social structures---they fundamentally reconstructed the fabric of Chinese society, from group identities to power relations.

To capture these transformations linguistically requires sources that span this historical period. We focus on analyzing official discourse through the People's Daily corpus and, secondarily, the Google Books Chinese corpus. These sources represent the only available large-scale longitudinal datasets covering this critical period in Chinese history. While this approach offers valuable insights, it necessitates careful interpretation. As the official newspaper of the Communist Party of China, the People's Daily represents authorized messaging rather than the full spectrum of social discourse. Similarly, published materials in China during this period underwent varying degrees of state influence. Therefore, our analysis primarily reveals how the Chinese state's representation of social groups evolved—representations that both reflected official ideology and exerted significant influence on public discourse.

Our methodological approach employs word embeddings at two temporal resolutions: decade-level analysis for identifying long-term evolutionary patterns, and annual resolution for capturing rapid semantic shifts and their correlation with specific historical events. This approach allows us to address three fundamental questions: (i) What are the overall trait representations of different social groups in Chinese state-sanctioned discourse across 70 years? (ii) How have official representations of social groups evolved during periods of rapid political and social transformation? (iii) How do these representations differ from those observed in Western societies?

Our findings demonstrate substantial divergence between stereotypes in Chinese state-controlled media and US corpora, particularly in representations of economic status, gender, and ethnicity. We find duality in evolution of social group representations in official discourse: while some groups demonstrate stability across revolutionary change, others undergo dramatic reversals during key periods. This work presents the first systematic analysis of how official linguistic representations evolved during revolutionary social change in a non-Western context and advances theoretical understanding of how state-sanctioned discourse encodes and maintains social structures across different ideological frameworks.

\section{Related Work}

\subsection{Computational analysis of social groups}

Word embeddings, which represent lexical items as dense vectors in high-dimensional space \citep{mikolov2013efficient,pennington2014glove}, have emerged as powerful analytical tools for quantitative social science \citep{kozlowski2019geometry}. These computational methods encode both explicit and implicit word relationships, proving particularly valuable for revealing nuanced shifts in societal attitudes and social group representations across different cultural and political contexts.

Research validating these methods for social analysis began with \citet{caliskan2017semantics}, who demonstrated that word embeddings trained on large corpora automatically encode human-like biases. Building on this foundation, \citet{garg2018word} leveraged historical word embeddings to track gender and ethnic stereotypes in the United States throughout the 20th century, demonstrating how computational tools can capture longitudinal sociological trends. \citet{charlesworth2023identifying,charlesworth2021gender,charlesworth2022historical} expanded this methodology by examining representations of diverse social groups across multiple corpora spanning 200 years, revealing evolutionary trajectories in trait content and latent structures across different social categories.

While these computational approaches have yielded valuable insights in Western contexts, particularly the United States, where researchers found gradual shifts in social attitudes and group representations, their application to societies undergoing rapid transformation, especially in non-Western contexts, remains limited. This gap is particularly significant because different social structures and historical trajectories may produce distinct patterns of change in group representation and valuation.

\subsection{Social change in modern China}

Since its establishment in 1949, the People's Republic of China has undergone one of the most dramatic instances of rapid societal change in modern history \citep{walder1989social}. As \citet{xie2011evidence} posits, the magnitude and velocity of China's modernization process may parallel other historical watersheds, such as the Industrial Revolution in 18th-century Britain.

This transformation unfolded through several pivotal periods that fundamentally restructured Chinese society. The Cultural Revolution (1966-1976) dramatically reordered social hierarchies, dissolving traditional status structures and creating new power dynamics. The subsequent Reform and Opening Up policy (1978) initiated market reforms that reshaped class relations and social mobility patterns. These changes profoundly impacted urban-rural relations, gender roles, and class structures \citep{jacka2014rural}, while the transition to a market economy generated novel social categories and redistributed power across society \citep{guthrie2012china}.

While qualitative research has documented these social transformations extensively, systematic quantitative analyses of their linguistic manifestations remain sparse. Recent developments in computational linguistics, particularly word embeddings, offer new opportunities to trace these changes through language patterns. Although some studies have applied computational methods to Chinese language \citep{hamamura2021individualism,zeng2015cultural,chen2022lexicon}, they have primarily focused on broad cultural values rather than specific social group representations. Analyzing how different social categories evolved through China's transformation using word embeddings can provide quantitative evidence for these historical shifts, offering new insights into how revolutionary social change shapes group identities and relationships.

\section{Methods}

\subsection{Data sources}
\begin{figure}[b!]
    \centering
    \includegraphics[width=\linewidth]{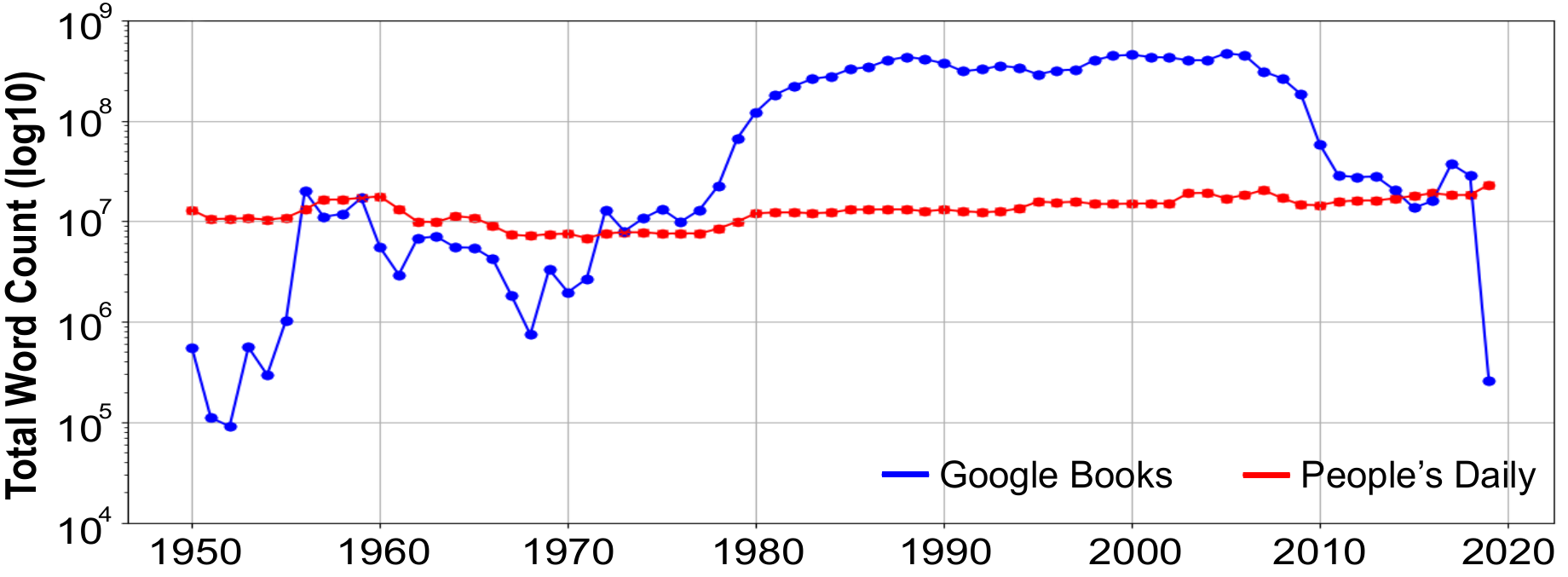}
    \caption{\textbf{Corpus size comparison across time.} We compare word counts between two Chinese language corpora from 1950 to 2019 The People's Daily corpus (red) maintains relatively stable coverage at approximately 10 million words per year throughout the period. The Google Books corpus (blue) shows temporal variation and particularly sparse coverage before 1980, with word counts fluctuating between 100 thousand and 100 million words per year.}
    \label{fig:data_source}
\end{figure}

Our analysis primarily relies on the People's Daily corpus~\citep{ps2023pdnd}, which contains approximately 2 million articles from China's official newspaper spanning 1946 to 2023. As the official organ of the Communist Party of China, this newspaper functioned as a key instrument of state communication rather than an independent journalistic entity. The discourse in this publication reflects official policy positions and approved ideological frameworks during each period studied. While this limits our ability to capture unofficial attitudes or dissenting views, it provides unique insight into how authorized social representations evolved within state-controlled discourse, which itself was a powerful force in shaping public understanding during this period.

The Google Books Chinese corpus, which we used as a secondary source, exhibits similar limitations. Published materials in China are subject to varying degrees of censorship and ideological control, particularly during politically sensitive periods. The temporal gaps in this corpus, especially during the 1950s-1970s, may reflect both practical constraints in data collection and the restricted publishing environment during certain historical periods. Due to space constraints, we primarily present findings from the People's Daily corpus in this paper, as it provides more consistent and comprehensive coverage throughout our study period; see \cref{fig:data_source}.

\subsection{Training word embeddings}

We employed the skip-gram \citep{mikolov2013efficient} with negative sampling model \citep{mikolov2013distributed} to train word embeddings for our diachronic analysis. For both corpora, we generated 300-dimensional word vectors with a context window of 3 words. We included words appearing more than 10 times in the People's Daily corpus, while the Google Books 5-grams inherently met this frequency threshold. To enable multi-scale temporal analysis, we trained embeddings at both decade-level intervals and annual resolutions from 1950 to 2019. The fine-grained annual embeddings allow us to detect subtle semantic shifts and rapid changes in social group representations that might be obscured at the coarser decade-level resolution.

\subsection{Selection and representation of social groups}

We selected social groups for analysis following four key criteria established by \citet{charlesworth2022historical}: (i) diversity in representation to capture a broad range of societal stereotypes in China, (ii) comprehensive coverage through multiple synonymous group labels (for example, representing ``\woman'' using formal terms like ``女性 (female)'', colloquial terms like ``姑娘 (young lady)'', and role-specific terms like ``妻子 (wife)''), (iii) clear comparison pairs (such as ``\woman'' \vs ``\man''), and (iv) semantic stability of group labels across time.

Our analysis examines stereotypes across five fundamental social dimensions; see \cref{tab:top10_traits}: gender, ethnicity, age, economic status, and body type. For the ethnic dimension, China's unique demographic context shaped our analytical approach---while \han{} comprise 91.5\% of the population, the remaining 8.5\% consists of 55 distinct ethnic minority groups \citep{jian2017recent}. Given this distribution, we analyze these diverse ethnic groups collectively under the category ``\minor'' rather than as individual ethnicities.

\subsection{Construction of a Chinese trait list with valence}

To analyze stereotypes systematically, we needed a comprehensive list of Chinese trait words with validated emotional valence scores. We built upon the work of \citet{xu2022valence}, who developed the largest available database of affective norms for simplified Chinese words, containing valence and arousal ratings for 11,310 words. Their database follows a methodology similar to the established Affective Norms for English Words and its expansions \citep{bradley1999affective,warriner2013norms}.

From this database, we extracted 465 trait-related words relevant to personality and character descriptions. However, applying contemporary valence ratings to historical text analysis requires careful consideration of potential semantic shifts. To address this, we validated the historical stability of each trait word's meaning and emotional connotation using ``Hanyu Da Cidian'' (汉语大词典), the most authoritative historical Chinese dictionary. This validation step ensures our valence-based analysis remains meaningful across different time periods.

\subsection{Computing social group representations from word embeddings}

To analyze stereotypes quantitatively, we first obtained reliable representations of how social groups associated with different traits in the embedding space. Following \citet{charlesworth2022historical} and \citet{manzini2019black}, we computed the Mean Average Cosine Similarity (MAC) between each social group's labels and trait words, accounting for temporal variation. For a social group $G$ with labels $L_{G,t} = \{l_1, ..., l_n\}$ at time period $t$ and a trait word $w$, the MAC score is:
\begin{equation}
    \text{MAC}(G, w, t) = \frac{1}{|L_{G,t}|} \sum_{l \in L_{G,t}} \cos(\vec{l_t}, \vec{w_t}),
\end{equation}
where $\vec{l_t}$ and $\vec{w_t}$ are the embedding vectors at time $t$, and $\cos(\cdot,\cdot)$ denotes cosine similarity. To ensure meaningful associations, we only consider traits where $\text{MAC}(G, w, t) \geq 0.2$.

Raw MAC scores indicate how strongly a group associates with particular traits. However, to understand the unique characteristics of each social group, we examine relative differences between comparison groups. We compute these differential trait associations as:
\begin{equation}
    \resizebox{0.9\hsize}{!}{$\displaystyle
        \text{DiffMAC}(G, C, w, t) = \text{MAC}(G, w, t) - \text{MAC}(C, w, t),%
    $}
\end{equation}
where $C$ represents the comparison group.

To identify traits that consistently distinguish groups across time, we aggregate these differences:
\begin{equation}
    \resizebox{0.9\hsize}{!}{$\displaystyle
        \text{AggDiffMAC}(G, C, w) = \frac{1}{|T_w|} \sum_{t \in T_w} \text{DiffMAC}(G, C, w, t),%
    $}
\end{equation}
where $T_w$ represents time periods where the trait exists in embeddings. We require traits to appear in at least 3 decades to ensure temporal stability. The traits with the highest positive AggDiffMAC scores represent the most distinctive and persistent characteristics associated with each group across time.

\subsection{Event-centric analysis using word embedding association test}

To analyze how historical events shaped social group representations over time, we developed an event-centric framework extending the \ac{weat} methodology \citep{caliskan2017semantics}. For each social group $G$, we compute a temporal \ac{weat} score measuring the differential association between positive and negative attributes:
\begin{equation}
    \text{\acs{weat}}(G, t) = \text{MAC}(G, P, t) - \text{MAC}(G, U, t),
\end{equation}
where $P$ and $U$ denote sets of pleasant and unpleasant attribute words at time period $t$. We adapted these attribute sets from \citet{caliskan2017semantics}, carefully translating them to Chinese while preserving cultural and linguistic nuances. To quantify the impact of specific historical events, we compared mean \ac{weat} scores within five-year windows before and after each event, assessing statistical significance through $t$-tests and effect sizes using Cohen's $d$.

\section{Results}

\subsection{Analysis 1: Systematic valence asymmetries in top 10 relative traits across 70 years}

\begin{table*}[ht!]
    \centering
    \scriptsize
    \setlength{\tabcolsep}{3pt}
    \caption{\textbf{Valence patterns in Chinese social group stereotypes (1950-2019).} Analysis of the top 10 traits reveals systematic biases: youth, thinness, ethnic minorities, and women are associated with superficially positive but often infantilizing traits (\eg, innocent, simple, pure), while elderly, wealthy, and overweight groups face predominantly negative characterizations. These trait associations suggest deep-rooted social biases encoded in language use across seven decades.}
    \label{tab:top10_traits}
    \begin{tabular}{ccc>{\tiny\arraybackslash}p{0.73\linewidth}}
        \toprule
        \textbf{Category} & \textbf{Group} & \textbf{Valence} & {\scriptsize\textbf{Top 10 Traits}} \\
        \midrule
        \multirow{2}{*}[-1.5ex]{Gender}  
        & woman & 1.24 & 天真无邪 (innocent), 恬静 (quiet), 神秘 (mysterious), 文静 (gentle), 优雅 (elegant), 活泼 (lively), 弱智 (stupid), 纯真 (pure), 快乐 (happy), 迷人 (charming) \\
        & man & -0.22 & 脚踏实地 (pragmatic), 决断 (decisive), 实干 (practical), 拖拉 (procrastinating), 不公 (unfair), 务实 (pragmatic), 墨守成规 (conservative), 自满 (complacent), 痴心妄想 (delusional), 自以为是 (self-righteous) \\
        \midrule
        \multirow{2}{*}[-1.5ex]{Ethnicity}
        & Han Chinese & -0.09 & 犹豫不决 (indecisive), 墨守成规 (conservative), 挑剔 (picky), 拖拉 (procrastinating), 随机应变 (adaptable), 武断 (arbitrary), 急躁 (impatient), 明智 (wise), 务实 (pragmatic), 别出心裁 (innovative) \\
        & ethnic minorities & 1.43 & 纯朴 (unsophisticated), 淳朴 (simple), 豪迈 (bold), 快乐 (happy), 活跃 (active), 创造力 (creative), 迷人 (charming), 阳光 (sunny), 脆弱 (fragile), 恬静 (quiet) \\
        \midrule
        \multirow{2}{*}[-1.5ex]{Age}
        & young & 1.37 & 优秀 (excellent), 刻苦 (hardworking), 朝气 (vigorous), 果敢 (resolute), 实干 (practical), 出众 (outstanding), 脚踏实地 (pragmatic), 凶狠 (fierce), 好胜 (competitive), 随机应变 (adaptable) \\
        & elderly & -1.44 & 痴呆 (senile), 麻木 (numb), 僵硬 (rigid), 奢侈 (luxurious), 愚昧 (ignorant), 漫不经心 (careless), 官僚 (bureaucratic), 消极 (negative), 悲观 (pessimistic), 烦躁 (irritable) \\
        \midrule
        \multirow{2}{*}[-1.5ex]{Economic}
        & rich & -1.53 & 官僚 (bureaucratic), 绅士 (gentleman), 独裁 (dictatorial), 贪得无厌 (greedy), 谄媚 (flattering), 背信弃义 (treacherous), 阿谀奉承 (sycophantic), 昏庸 (incompetent), 专横 (arbitrary), 吹嘘 (boastful) \\
        & poor & 0.41 & 弱智 (foolish), 痴呆 (muddled) 刻苦 (hardworking), 落后 (underdeveloped), 体贴 (considerate), 安静 (quiet), 薄弱 (weak), 关心 (caring), 体贴入微 (thoughtful), 无私 (selfless) \\
        \midrule
        \multirow{2}{*}{Body type}
        & fat & -1.40 & 不良 (bad), 无害 (harmless), 负面 (negative), 极端 (extreme), 恐怖 (terrible), 排外 (xenophobic), 盲目 (blind), 迷信 (superstitious), 弱智 (stupid), 不切实际 (impractical) \\
        & thin & 1.41 & 骁勇 (valiant), 端庄 (dignified), 坚毅 (persevering), 刚毅 (resolute), 刚强 (strong), 倔强 (stubborn), 妩媚 (charming), 雍容 (graceful), 耿直 (honest), 豪爽 (generous) \\
        \bottomrule
    \end{tabular}
\end{table*}

Using decade-level embeddings spanning 70 years, we analyzed AggDiffMAC scores to identify persistent trait representations for each social group; see \cref{tab:top10_traits}. The most pronounced valence disparities emerged in age and body type dimensions (both $2.81$). Young ($1.37$) and Thin ($1.41$) groups were characterized by traits suggesting competence and strength (\eg, excellent, hardworking, valiant), while Elderly ($-1.44$) and Fat ($-1.40$) groups were associated with decline and deficiency (\eg, senile, numb, bad).

Economic status showed the second-largest valence gap ($1.94$). Rich received strongly negative valence ($-1.53$), with traits suggesting corruption and power abuse (\eg, bureaucratic, dictatorial, treacherous). In contrast, Poor showed slightly positive valence ($0.41$), reflecting both virtuous qualities (\eg, hardworking, caring, selfless) and diminished capability (\eg, foolish, muddled, weak).

Ethnic ($1.52$) and gender ($1.46$) representations exhibited smaller but significant disparities. Ethnic minorities received high positive valence ($1.43$) emphasizing simplicity and vibrancy (\eg, simple, bold, happy), while Han Chinese showed near-neutral valence ($-0.09$) with pragmatic associations (\eg, indecisive, conservative, pragmatic). Gender analysis revealed women with positive valence ($1.24$) characterized by passive traits (\eg, innocent, quiet, gentle), while men showed slightly negative valence ($-0.22$) with action-oriented traits (\eg, pragmatic, decisive, practical).

These systematic asymmetries reveal persistent biases in Chinese linguistic representations. Importantly, even positive valence scores often masked problematic stereotypes, such as the infantilization of women and ethnic minorities through traits emphasizing innocence and simplicity. The consistent negative portrayal of elderly, fat, and wealthy groups, alongside patronizing positive portrayals of other groups, suggests deeply embedded social biases in linguistic patterns.

\subsection{Analysis 2: Temporal evolution of social group representations}

Tracking valence asymmetries from 1950 to 2019 through decade-wise DiffMAC scores revealed two distinct patterns in social group representations: stable hierarchical patterns and dramatic reversals.

\paragraph*{Stable hierarchical patterns}

\begin{figure}[b!]
    \centering
    \begin{subfigure}[b]{0.5\linewidth}
        \centering
        \includegraphics[width=\linewidth]{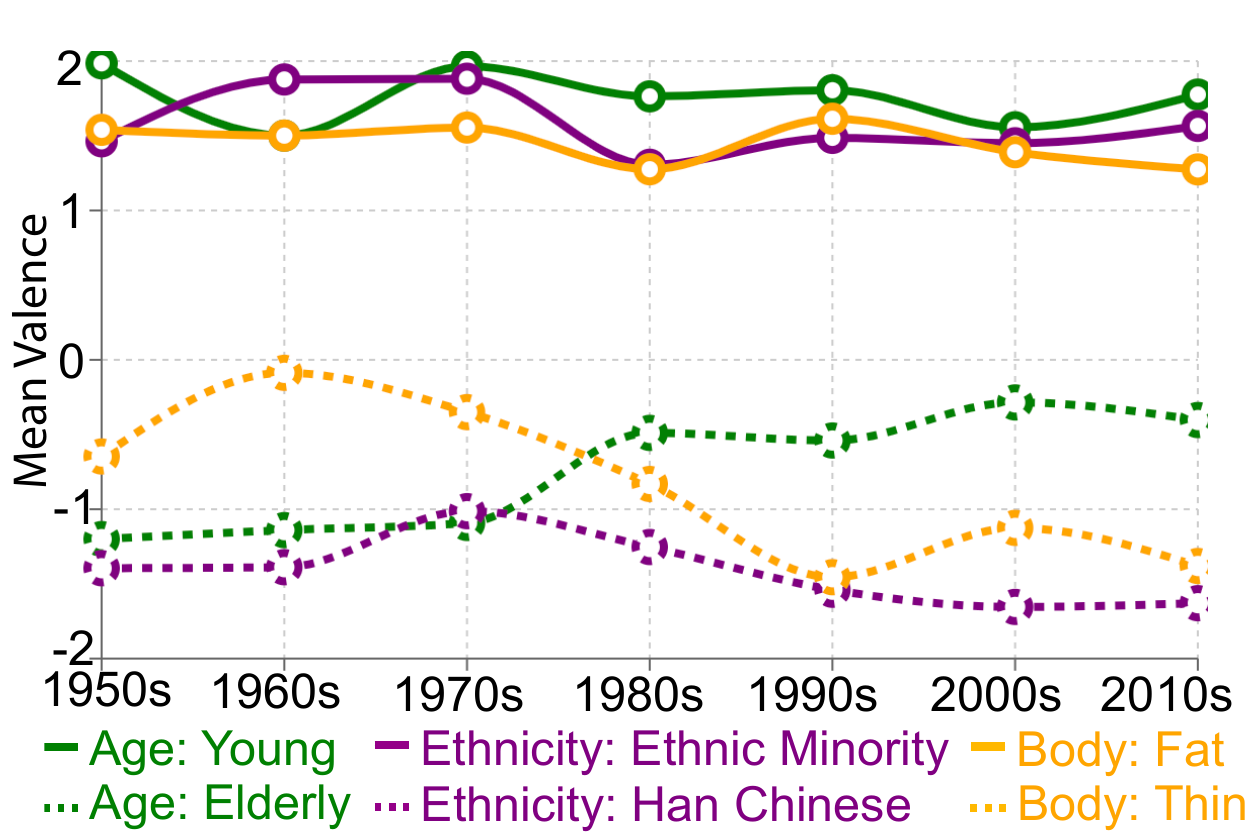}
        \caption{stable hierarchical patterns}
        \label{fig:stable}
    \end{subfigure}%
    \begin{subfigure}[b]{0.5\linewidth}
        \centering
        \includegraphics[width=\linewidth]{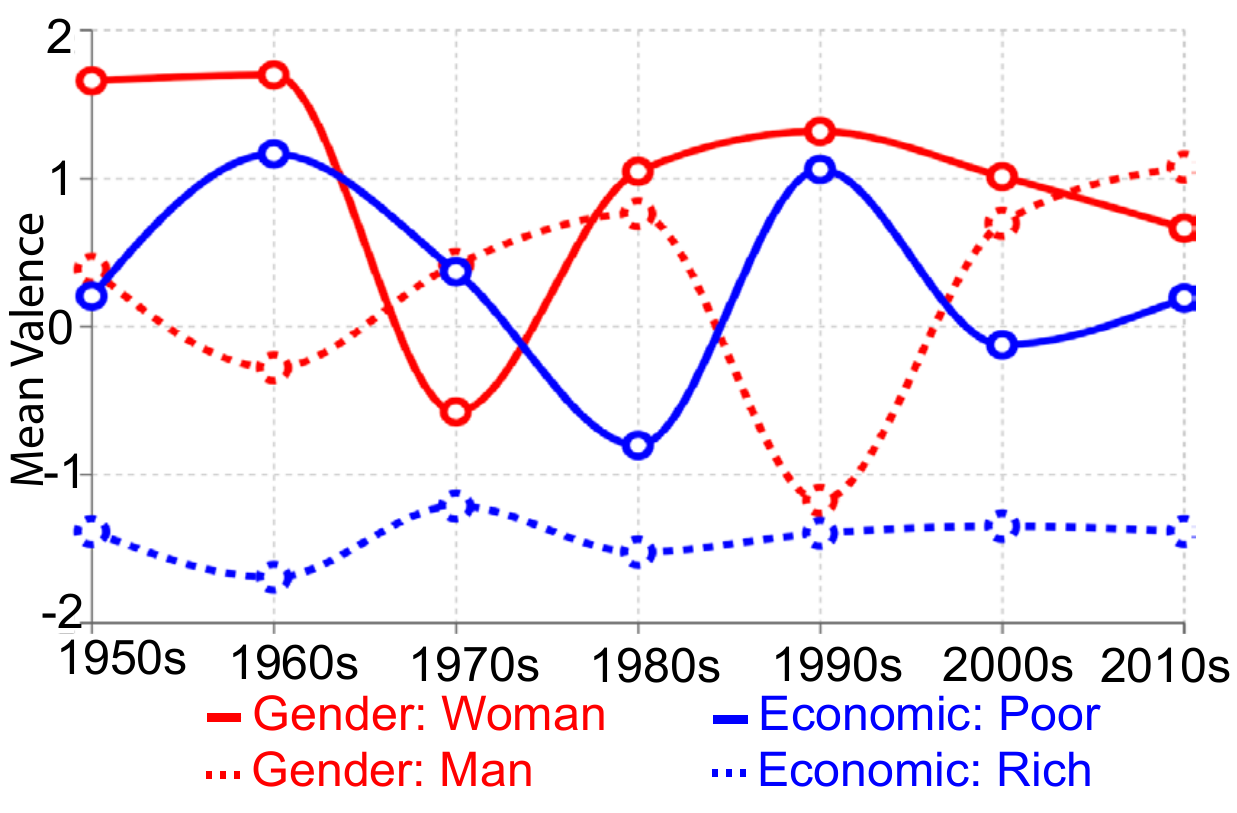}
        \caption{reversals/sharp transitions}
        \label{fig:shift}
    \end{subfigure}%
    \caption{\textbf{Temporal evolution of Chinese social group representations (1950-2019).} (a) Mean valence scores show persistent patterns across decades, with consistently higher valence for young (\vs elderly), ethnic minority (\vs Han Chinese), and thin (\vs fat) groups. (b) Mean valence scores demonstrate dramatic shifts and reversals for gender and economic status groups, including the 1970s gender valence reversal and the post-1978 transition in poor-rich dynamics.}
\end{figure}

The ethnic minority-majority dynamic showed exceptional stability; see visualizations in \cref{fig:stable}. Ethnic minorities maintained consistently positive associations (mean valence: $1.31$ to $1.88$), with ``淳朴 (simple)'' and ``纯朴 (unsophisticated)'' appearing as defining traits across all decades, followed by ``快乐/喜乐(happy)''. Han Chinese representations remained predominantly negative (mean valence: $-1.01$ to $-1.65$), sustaining a valence gap of $2.56$ to $3.27$.

Age-based representations displayed similar hierarchical persistence. Young maintained positive valence ($1.50$ to $1.97$), while Elderly stayed negative ($-0.28$ to $-1.20$). The intergenerational gap peaked in the 1950s ($3.19$) before gradually narrowing, primarily through reduced negative elderly associations rather than shifts in youth representations.

Body type representations exhibited comparable consistency. Thin retained positive associations ($1.28$ to $1.61$), while Fat faced persistent negative characterization ($-0.35$ to $-1.38$). This enduring valence disparity reflects deeply embedded societal standards regarding physical appearance.

\subsubsection{Pattern reversals and sharp transitions}

The representation of women exhibited marked volatility; see the visualization in \cref{fig:shift}. Strong positive valence in the 1950s-60s ($1.66$ to $1.70$) emphasized traits like ``活泼 (lively),'' ``恬静 (serene),'' and ``天真无邪 (innocent).'' This pattern reversed sharply in the 1970s (see \cref{tab:women-traits}) as valence plummeted to $-0.57$, with emergent negative traits like ``黑心 (evil)'' and ``无耻 (shameless).'' While recovering in the 1980s ($1.05$) and peaking in the 1990s ($1.32$), female valence gradually declined to $0.66$ by the 2010s. Male representations showed parallel fluctuations, from positive in the 1950s ($0.39$) to negative in the 1960s ($-0.28$), reaching their nadir in the 1990s ($-1.17$) before recovering strongly. The 2010s marked a historic reversal as male valence ($1.08$) surpassed female valence ($0.66$) for the first time outside the 1970s anomaly, characterized by traits like ``大义 (righteousness)'' and ``正气凛然 (upright and noble).''

\begin{table}[ht!]
    \scriptsize
    \centering
    \caption{\textbf{Top 10 traits most strongly associated with women (1960s--1980s).} Each column shows distinctive trait patterns with corresponding mean valence scores: predominantly positive traits in the 1960s ($1.70$), shifting to negative traits in the 1970s ($-0.57$), before returning to positive traits in the 1980s ($1.05$).}
    \label{tab:women-traits}
    \begin{tabular}{lll}
        \toprule
        \textbf{1960s ($1.70$)} & \textbf{1970s ($-0.57$)} & \textbf{1980s ($1.05$)}\\
        \midrule
        懂事 (sensible) & 忠孝 (filial) & 纯真 (naive) \\
        可爱 (cute) & 忠义 (loyal) & 文静 (gentle) \\
        淳朴 (simple) & 黑心 (evil) & 腼腆 (shy) \\
        天真无邪 (innocent) & 滑稽 (ridiculous) & 娇生惯养 (pampered) \\
        善良 (kind) & 无耻 (shameless) & 恬静 (composed) \\
        多情 (affectionate) & 颓废 (decadent) & 淘气 (naughty) \\
        幽默 (humorous) & 污秽 (filthy) & 端庄 (dignified) \\
        活泼 (lively) & 中庸 (mediocre) & 优雅 (elegant) \\
        纯真 (naive) & 正气凛然 (righteous) & 迷人 (charming) \\
        快乐 (happy) & 下流 (vulgar) & 天真无邪 (innocent) \\
        \bottomrule
    \end{tabular}
\end{table}

Economic status representations showed another clear transition. While Rich maintained consistent negative valence ($-1.21$ to $-1.69$), Poor experienced a dramatic shift from positive ($1.17$) to negative ($-0.80$) valence in the 1980s, coinciding with China's post-1978 economic reforms.

\paragraph{Summary}

These temporal dynamics reveal distinct patterns of stability and change in social representations across modern China. While representations of ethnicity, age, and body type maintained consistent hierarchical structures, gender and economic status underwent dramatic reorganization during periods of social and political transformation. This heterogeneity suggests that different dimensions of social hierarchy exhibit varying degrees of resilience to broader societal changes, reflecting the complex interplay between linguistic representation and social evolution.

\subsection{Analysis 3: Quantifying social change through year-to-year embeddings}

To analyze social change dynamics in depth, we examined trait-group semantic associations from 1950-2019 using (i) temporal correlation analysis to identify the pace of change, and (ii) cross-sectional analysis during major historical events to capture differential impacts across social groups. These analyses relied on embeddings trained separately for each year, allowing us to track fine-grained semantic shifts.

\begin{figure}[t!]
    \centering
    \begin{subfigure}[b]{0.5\linewidth}
        \centering
        \includegraphics[width=\linewidth]{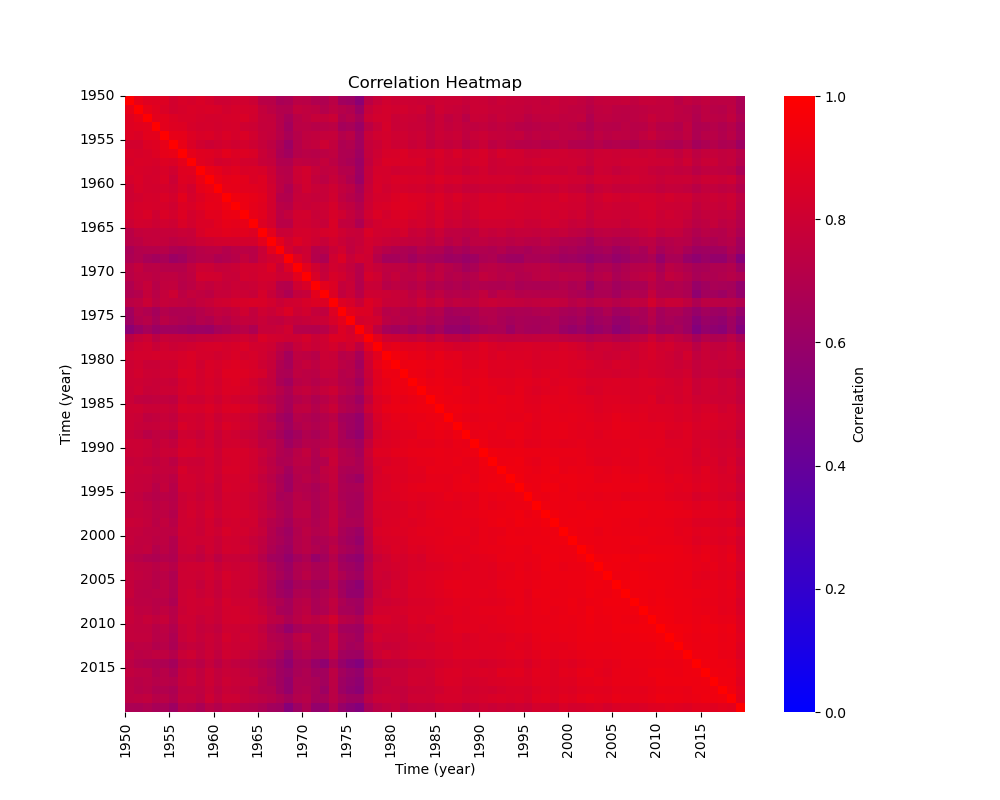}
        \caption{woman}
    \end{subfigure}%
    \begin{subfigure}[b]{0.5\linewidth}
        \centering
        \includegraphics[width=\linewidth]{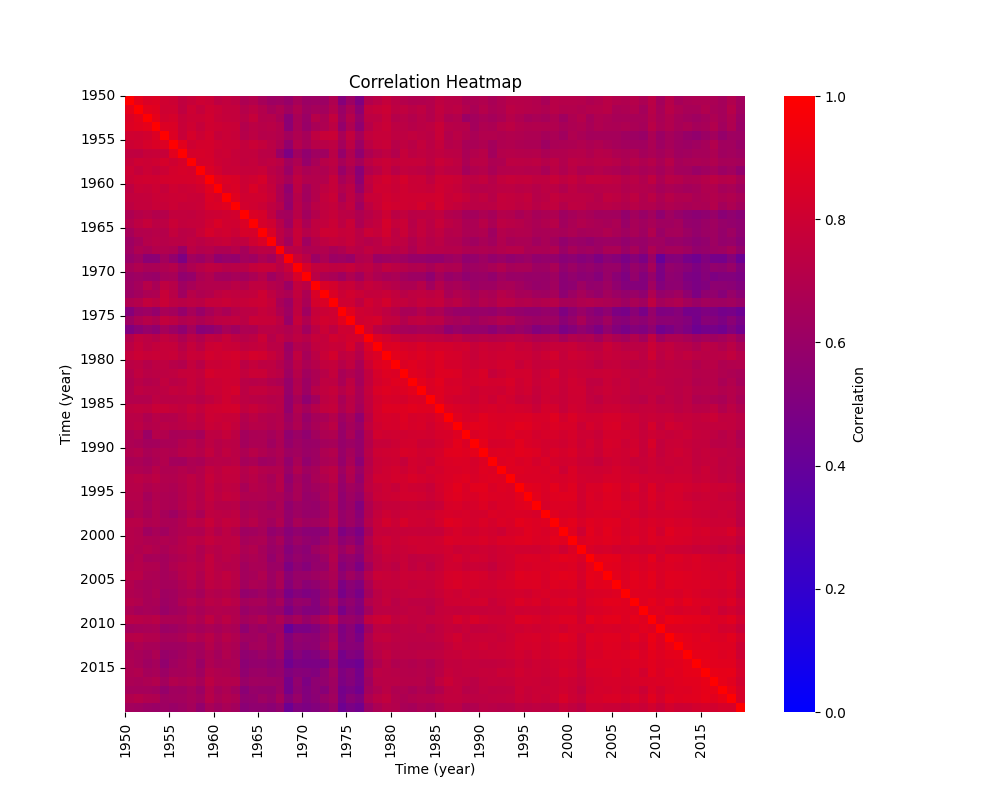}
        \caption{young}
    \end{subfigure}%
    \\%
    \begin{subfigure}[b]{0.5\linewidth}
        \centering
        \includegraphics[width=\linewidth]{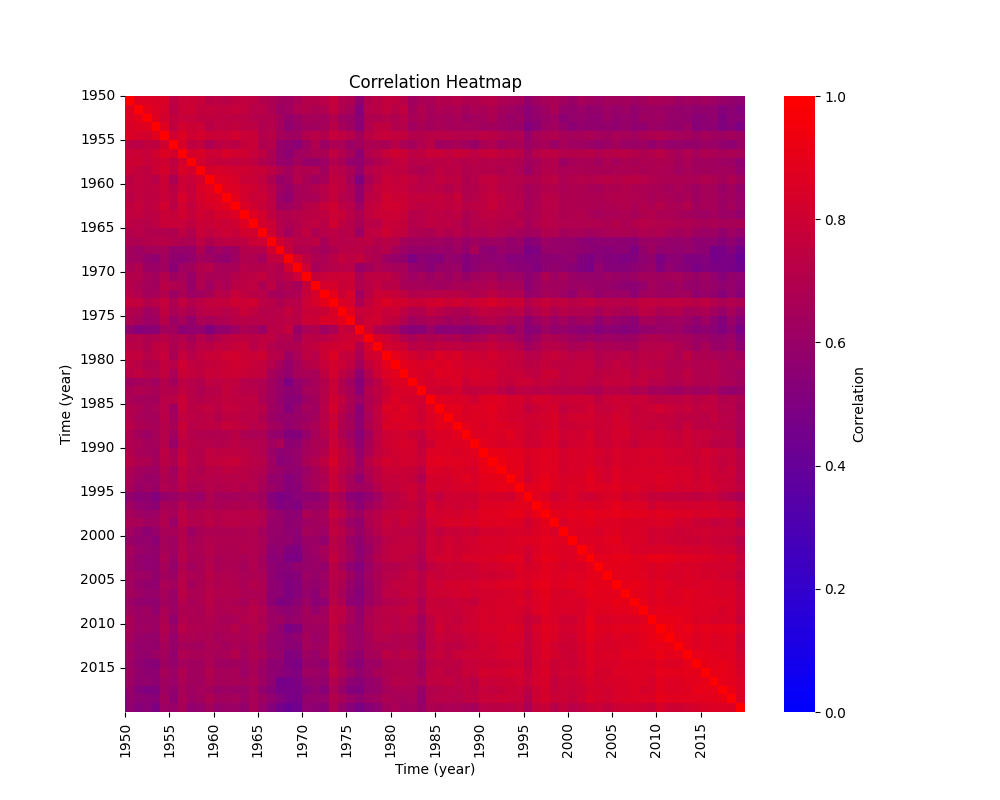}
        \caption{poor}
    \end{subfigure}%
    \begin{subfigure}[b]{0.5\linewidth}
        \centering
        \includegraphics[width=\linewidth]{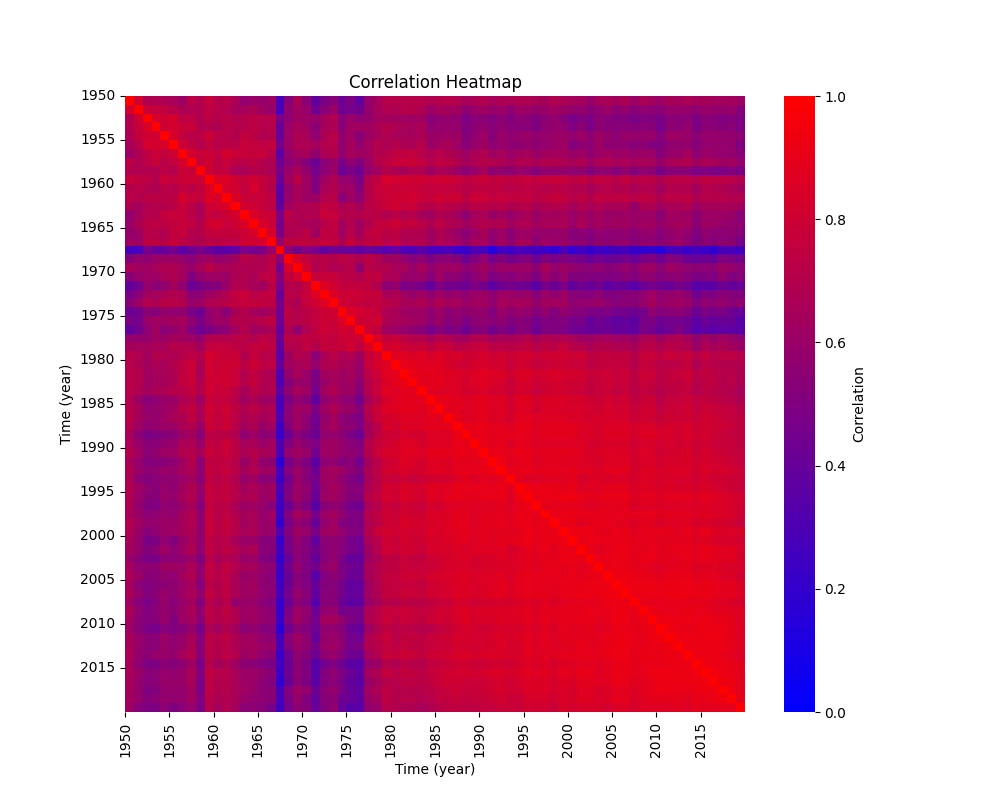}
        \caption{ethnic minorities}
    \end{subfigure}%
    \caption{\textbf{Year-to-year correlation matrices revealing temporal dynamics in social representations (1950-2019).} Each heatmap displays correlation coefficients between trait associations across years (red: strong positive; blue: weaker correlations; range: $0$ t0 $1.0$). Notable observations include (i) strong year-to-year stability near the diagonal ($r>0.8$); (ii) a distinct band of lower correlations during 1966-1976 (r<0.4), particularly evident in the ``Young'' and ``Minorities'' matrices; and (iii) varying impacts across groups during this period, with ``Woman'' and ``Poor'' showing relatively higher stability. Since paired groups under the same social category demonstrated similar correlation patterns, we present one representative plot from each category. (Vector graphics; zoom for details.)}
    \label{fig:combined-heatmaps}
\end{figure}

\paragraph*{Temporal stability and evolution}

Our correlation analysis uncovered two fundamental patterns in how social representations evolved. First, semantic associations demonstrated remarkable stability over short time spans, with correlation coefficients above $0.8$ between consecutive years. This high year-to-year stability suggests that social representations typically resist rapid changes under normal circumstances. Second, these correlations weakened systematically as temporal distance grew, declining to moderate levels (between $0.4$ and $0.6$) across multi-decade intervals. Rather than sudden shifts, this pattern reveals a gradual evolution of social representations over longer timescales, where small annual changes accumulated into substantial transformations across decades.

The period of 1966-1976, coinciding with China's Cultural Revolution, emerged as a striking exception to these patterns. During this decade, correlations dropped dramatically ($r<0.4$) compared to all other periods, visible as a distinctive band of reduced correlations across all eight matrices; see \cref{fig:combined-heatmaps}. This unprecedented disruption in semantic stability indicates that the Cultural Revolution catalyzed rapid changes in how social groups were characterized and perceived.

\subsubsection{Event-centric analysis of social change}

Building on these correlation patterns, we conducted detailed analyses of two transformative periods: the Cultural Revolution (1966-1976) and the Economic Reform era (post-1978). The Cultural Revolution triggered immediate and divergent impacts across social groups. Poor experienced a significant positive shift in \ac{weat} scores ($t(67)=2.768, p< .01, d=0.682$), while both Woman ($t(142)=-4.369, p< .001, d=-0.740$) and Young ($t(49)=-2.352, p< .05, d=-0.672$) saw significant declines. While our decade-level analysis captured the overall trajectory of these changes, these fine-grained measurements reveal how rapidly social representations could transform in response to political upheaval. 

The Economic Reform period brought its own distinctive shifts. Rich showed significantly improved \ac{weat} scores ($t(74)=3.508, p< .001, d=0.817$), suggesting that economic liberalization weakened previously negative associations with wealth. However, this shift was partial---trait associations with the rich maintained an overall negative valence throughout this period, highlighting the persistence of certain social attitudes even amid the economic transformation.

\section{Discussions}

Our analysis of representations in Chinese state-controlled media from 1950 to 2019 illuminates both convergent patterns shared with Western contexts and distinctive characteristics shaped by China's unique sociopolitical environment. These findings address our three central research questions while revealing complex interactions between official discourse, state ideology, and social transformation.

\subsection{Mechanisms of social change in China}

Addressing our first research question on overall trait representations across 70 years, our word embedding analysis reveals how social representations evolved in state-controlled media. Different social categories exhibit varying resilience to officially sanctioned transformation. While age, ethnicity, and body type maintained relative stability even during upheavals, gender and economic status proved more susceptible to political intervention. This asymmetry suggests that state influence interacts unevenly with existing cultural schemas. 

For instance, the Cultural Revolution rapidly reshaped gender representations through propaganda campaigns, yet traces of traditional gender roles re-emerged in later decades. Similarly, while wealth was vilified under Maoist ideology, negative associations with affluence persisted even after pro-market reforms, reflecting deeper moral narratives about labor and fairness. These patterns highlight how social representations are shaped not only by top-down agendas but also by enduring cultural logics, echoing Bourdieu's view of symbolic power as a negotiation between institutional force and collective disposition~\citep{bourdieu1979symbolic}.

\subsection{Temporal dynamics of group-trait associations}

In response to our second research question on how official representations evolved during periods of rapid transformation, our analysis reveals distinct temporal patterns in state discourse. Both Chinese and English contexts exhibit gradual correlation decay over time, but China's Cultural Revolution (1966-1976) stands as an exception. During this period, correlations in official discourse plummeted ($r<0.4$) relative to other decades, manifesting as a band of reduced correlations---evidence that state-directed interventions can disrupt semantic continuity. This discontinuity in semantic space demonstrates how revolutionary periods create linguistic ruptures rather than evolutionary shifts. This finding carries methodological implications for diachronic language analysis. Future studies must consider both the temporal distance between measurements and the impact of major historical events that accelerate semantic change, advancing our understanding of how sociopolitical forces interact with semantic evolution in transitional societies.

\subsection{Cross-cultural patterns in social group representations}

Addressing our third question comparing Chinese state-sanctioned discourse with Western representations, we find both similarities and critical differences. Western contexts associate women with personality traits and men with achievement \citep{charlesworth2021gender, charlesworth2022historical, charlesworth2023identifying}. Chinese state media shows similar benevolent sexism toward women through traits emphasizing passivity—superficially positive characterizations that mask restrictive stereotypes. Men's representation reveals a notable contrast—Chinese state media portrays more negative masculine traits (\eg, ``不公(unfair)'', ``自满(complacent)'') which paradoxically encode privilege by implying authority and freedom to display flaws while maintaining status. 

Political ideology shapes economic and ethnic representations distinctively. Unlike positive U.S. portrayals \citep{charlesworth2022historical, charlesworth2023identifying}, Chinese state discourse represents the Rich negatively, reflecting socialist ideology's persistence despite market reforms. Ethnic representations also differ: U.S. media shows negative associations with minorities \citep{charlesworth2022historical}, while Chinese state media employs ``positive othering~\citep{rosello1998declining}'' through traits suggesting primitive simplicity (``纯朴(unsophisticated),'' ``淳朴(simple)''). These patterns reveal how state-controlled media in China reflects both universal cognitive tendencies and specific ideological frameworks, demonstrating distinct mechanisms by which different systems maintain social hierarchies.

\subsection{Limitations and future directions}

Several limitations warrant consideration. Our reliance on the People's Daily, a state-controlled newspaper, may not capture the full spectrum of everyday social discourse, particularly dissenting views and informal language. The application of contemporary valence ratings to historical text may not fully account for semantic shifts over time. Finally, while our computational approach enables large-scale analysis, it may miss nuanced contextual meanings and cultural connotations. Future research could address these limitations by incorporating diverse textual sources, developing historical valence measures, and conducting more fine-grained analyses of social group representations across different contexts.

\section{Acknowledgments}

We gratefully acknowledge Siyuan Yin, Haotian Zhang, Xingyu Liu, and Boyuan Chen for their valuable assistance with data exploration, and Guangyuan Jiang for his insightful guidance on research framing. This work is supported in part by the State Key Lab of General AI at Peking University, the PKU-BingJi Joint Laboratory for Artificial Intelligence, and the National Comprehensive Experimental Base for Governance of Intelligent Society, Wuhan East Lake High-Tech Development Zone.

\bibliographystyle{apacite}
\balance
\renewcommand\bibliographytypesize{\small}
\setlength{\bibleftmargin}{.125in}
\setlength{\bibindent}{-\bibleftmargin}
\bibliography{reference_header,reference}

\begin{thebibliography}{}

\bibitem [\protect \citeauthoryear {%
Bourdieu%
}{%
Bourdieu%
}{%
{\protect \APACyear {1979}}%
}]{%
bourdieu1979symbolic}
\APACinsertmetastar {%
bourdieu1979symbolic}%
\begin{APACrefauthors}%
Bourdieu, P.%
\end{APACrefauthors}%
\unskip\
\newblock
\APACrefYearMonthDay{1979}{}{}.
\newblock
{\BBOQ}\APACrefatitle {Symbolic power} {Symbolic power}.{\BBCQ}
\newblock
\APACjournalVolNumPages{Critique of anthropology}{4}{13-14}{77--85}.
\PrintBackRefs{\CurrentBib}

\bibitem [\protect \citeauthoryear {%
Bourdieu%
}{%
Bourdieu%
}{%
{\protect \APACyear {1991}}%
}]{%
bourdieu1991language}
\APACinsertmetastar {%
bourdieu1991language}%
\begin{APACrefauthors}%
Bourdieu, P.%
\end{APACrefauthors}%
\unskip\
\newblock
\APACrefYearMonthDay{1991}{}{}.
\newblock
{\BBOQ}\APACrefatitle {Language and symbolic power} {Language and symbolic
  power}.{\BBCQ}
\newblock
\APACjournalVolNumPages{Polity}{}{}{}.
\PrintBackRefs{\CurrentBib}

\bibitem [\protect \citeauthoryear {%
Bradley%
}{%
Bradley%
}{%
{\protect \APACyear {1999}}%
}]{%
bradley1999affective}
\APACinsertmetastar {%
bradley1999affective}%
\begin{APACrefauthors}%
Bradley, M.%
\end{APACrefauthors}%
\unskip\
\newblock
\APACrefYearMonthDay{1999}{}{}.
\newblock
{\BBOQ}\APACrefatitle {Affective Norms for English Words (ANEW): Instruction
  Manual and Affective Ratings} {Affective norms for english words (anew):
  Instruction manual and affective ratings}.{\BBCQ}
\newblock
\APACjournalVolNumPages{Technical Report C-1}{}{}{}.
\PrintBackRefs{\CurrentBib}

\bibitem [\protect \citeauthoryear {%
Caliskan%
, Bryson%
\BCBL {}\ \BBA {} Narayanan%
}{%
Caliskan%
\ \protect \BOthers {.}}{%
{\protect \APACyear {2017}}%
}]{%
caliskan2017semantics}
\APACinsertmetastar {%
caliskan2017semantics}%
\begin{APACrefauthors}%
Caliskan, A.%
, Bryson, J\BPBI J.%
\BCBL {}\ \BBA {} Narayanan, A.%
\end{APACrefauthors}%
\unskip\
\newblock
\APACrefYearMonthDay{2017}{}{}.
\newblock
{\BBOQ}\APACrefatitle {Semantics derived automatically from language corpora
  contain human-like biases} {Semantics derived automatically from language
  corpora contain human-like biases}.{\BBCQ}
\newblock
\APACjournalVolNumPages{Science}{356}{6334}{183--186}.
\PrintBackRefs{\CurrentBib}

\bibitem [\protect \citeauthoryear {%
Cameron%
}{%
Cameron%
}{%
{\protect \APACyear {2012}}%
}]{%
cameron2012verbal}
\APACinsertmetastar {%
cameron2012verbal}%
\begin{APACrefauthors}%
Cameron, D.%
\end{APACrefauthors}%
\unskip\
\newblock
\APACrefYear{2012}.
\newblock
\APACrefbtitle {Verbal hygiene} {Verbal hygiene}.
\newblock
\APACaddressPublisher{}{Routledge}.
\PrintBackRefs{\CurrentBib}

\bibitem [\protect \citeauthoryear {%
Charlesworth%
, Caliskan%
\BCBL {}\ \BBA {} Banaji%
}{%
Charlesworth%
\ \protect \BOthers {.}}{%
{\protect \APACyear {2022}}%
}]{%
charlesworth2022historical}
\APACinsertmetastar {%
charlesworth2022historical}%
\begin{APACrefauthors}%
Charlesworth, T\BPBI E.%
, Caliskan, A.%
\BCBL {}\ \BBA {} Banaji, M\BPBI R.%
\end{APACrefauthors}%
\unskip\
\newblock
\APACrefYearMonthDay{2022}{}{}.
\newblock
{\BBOQ}\APACrefatitle {Historical representations of social groups across 200
  years of word embeddings from Google Books} {Historical representations of
  social groups across 200 years of word embeddings from google books}.{\BBCQ}
\newblock
\APACjournalVolNumPages{{Proceedings of the National Academy of Sciences
  (PNAS)}}{119}{28}{e2121798119}.
\PrintBackRefs{\CurrentBib}

\bibitem [\protect \citeauthoryear {%
Charlesworth%
, Sanjeev%
, Hatzenbuehler%
\BCBL {}\ \BBA {} Banaji%
}{%
Charlesworth%
\ \protect \BOthers {.}}{%
{\protect \APACyear {2023}}%
}]{%
charlesworth2023identifying}
\APACinsertmetastar {%
charlesworth2023identifying}%
\begin{APACrefauthors}%
Charlesworth, T\BPBI E.%
, Sanjeev, N.%
, Hatzenbuehler, M\BPBI L.%
\BCBL {}\ \BBA {} Banaji, M\BPBI R.%
\end{APACrefauthors}%
\unskip\
\newblock
\APACrefYearMonthDay{2023}{}{}.
\newblock
{\BBOQ}\APACrefatitle {Identifying and predicting stereotype change in large
  language corpora: 72 groups, 115 years (1900--2015), and four text sources}
  {Identifying and predicting stereotype change in large language corpora: 72
  groups, 115 years (1900--2015), and four text sources}.{\BBCQ}
\newblock
\APACjournalVolNumPages{Journal of Personality and Social Psychology}{}{}{}.
\PrintBackRefs{\CurrentBib}

\bibitem [\protect \citeauthoryear {%
Charlesworth%
, Yang%
, Mann%
, Kurdi%
\BCBL {}\ \BBA {} Banaji%
}{%
Charlesworth%
\ \protect \BOthers {.}}{%
{\protect \APACyear {2021}}%
}]{%
charlesworth2021gender}
\APACinsertmetastar {%
charlesworth2021gender}%
\begin{APACrefauthors}%
Charlesworth, T\BPBI E.%
, Yang, V.%
, Mann, T\BPBI C.%
, Kurdi, B.%
\BCBL {}\ \BBA {} Banaji, M\BPBI R.%
\end{APACrefauthors}%
\unskip\
\newblock
\APACrefYearMonthDay{2021}{}{}.
\newblock
{\BBOQ}\APACrefatitle {Gender stereotypes in natural language: Word embeddings
  show robust consistency across child and adult language corpora of more than
  65 million words} {Gender stereotypes in natural language: Word embeddings
  show robust consistency across child and adult language corpora of more than
  65 million words}.{\BBCQ}
\newblock
\APACjournalVolNumPages{Psychological Science}{32}{2}{218--240}.
\PrintBackRefs{\CurrentBib}

\bibitem [\protect \citeauthoryear {%
Chen%
, Chersoni%
\BCBL {}\ \BBA {} Huang%
}{%
Chen%
\ \protect \BOthers {.}}{%
{\protect \APACyear {2022}}%
}]{%
chen2022lexicon}
\APACinsertmetastar {%
chen2022lexicon}%
\begin{APACrefauthors}%
Chen, J.%
, Chersoni, E.%
\BCBL {}\ \BBA {} Huang, C\BHBI r.%
\end{APACrefauthors}%
\unskip\
\newblock
\APACrefYearMonthDay{2022}{}{}.
\newblock
{\BBOQ}\APACrefatitle {Lexicon of changes: towards the evaluation of diachronic
  semantic shift in Chinese} {Lexicon of changes: towards the evaluation of
  diachronic semantic shift in chinese}.{\BBCQ}
\newblock
\BIn{} \APACrefbtitle {Workshop on Computational Approaches to Historical
  Language Change.} {Workshop on computational approaches to historical
  language change.}
\PrintBackRefs{\CurrentBib}

\bibitem [\protect \citeauthoryear {%
Fairclough%
}{%
Fairclough%
}{%
{\protect \APACyear {2013}}%
}]{%
fairclough2013language}
\APACinsertmetastar {%
fairclough2013language}%
\begin{APACrefauthors}%
Fairclough, N.%
\end{APACrefauthors}%
\unskip\
\newblock
\APACrefYear{2013}.
\newblock
\APACrefbtitle {Language and power} {Language and power}.
\newblock
\APACaddressPublisher{}{Routledge}.
\PrintBackRefs{\CurrentBib}

\bibitem [\protect \citeauthoryear {%
Garg%
, Schiebinger%
, Jurafsky%
\BCBL {}\ \BBA {} Zou%
}{%
Garg%
\ \protect \BOthers {.}}{%
{\protect \APACyear {2018}}%
}]{%
garg2018word}
\APACinsertmetastar {%
garg2018word}%
\begin{APACrefauthors}%
Garg, N.%
, Schiebinger, L.%
, Jurafsky, D.%
\BCBL {}\ \BBA {} Zou, J.%
\end{APACrefauthors}%
\unskip\
\newblock
\APACrefYearMonthDay{2018}{}{}.
\newblock
{\BBOQ}\APACrefatitle {Word embeddings quantify 100 years of gender and ethnic
  stereotypes} {Word embeddings quantify 100 years of gender and ethnic
  stereotypes}.{\BBCQ}
\newblock
\APACjournalVolNumPages{{Proceedings of the National Academy of Sciences
  (PNAS)}}{115}{16}{E3635--E3644}.
\PrintBackRefs{\CurrentBib}

\bibitem [\protect \citeauthoryear {%
Guthrie%
}{%
Guthrie%
}{%
{\protect \APACyear {2012}}%
}]{%
guthrie2012china}
\APACinsertmetastar {%
guthrie2012china}%
\begin{APACrefauthors}%
Guthrie, D.%
\end{APACrefauthors}%
\unskip\
\newblock
\APACrefYear{2012}.
\newblock
\APACrefbtitle {China and globalization: The social, economic and political
  transformation of Chinese society} {China and globalization: The social,
  economic and political transformation of chinese society}.
\newblock
\APACaddressPublisher{}{Routledge}.
\PrintBackRefs{\CurrentBib}

\bibitem [\protect \citeauthoryear {%
Hamamura%
, Chen%
, Chan%
, Chen%
\BCBL {}\ \BBA {} Kobayashi%
}{%
Hamamura%
\ \protect \BOthers {.}}{%
{\protect \APACyear {2021}}%
}]{%
hamamura2021individualism}
\APACinsertmetastar {%
hamamura2021individualism}%
\begin{APACrefauthors}%
Hamamura, T.%
, Chen, Z.%
, Chan, C\BPBI S.%
, Chen, S\BPBI X.%
\BCBL {}\ \BBA {} Kobayashi, T.%
\end{APACrefauthors}%
\unskip\
\newblock
\APACrefYearMonthDay{2021}{}{}.
\newblock
{\BBOQ}\APACrefatitle {Individualism with Chinese characteristics? Discerning
  cultural shifts in China using 50 years of printed texts.} {Individualism
  with chinese characteristics? discerning cultural shifts in china using 50
  years of printed texts.}{\BBCQ}
\newblock
\APACjournalVolNumPages{American Psychologist}{76}{6}{888}.
\PrintBackRefs{\CurrentBib}

\bibitem [\protect \citeauthoryear {%
Jacka%
}{%
Jacka%
}{%
{\protect \APACyear {2014}}%
}]{%
jacka2014rural}
\APACinsertmetastar {%
jacka2014rural}%
\begin{APACrefauthors}%
Jacka, T.%
\end{APACrefauthors}%
\unskip\
\newblock
\APACrefYear{2014}.
\newblock
\APACrefbtitle {Rural women in urban China: Gender, migration, and social
  change} {Rural women in urban china: Gender, migration, and social change}.
\newblock
\APACaddressPublisher{}{Routledge}.
\PrintBackRefs{\CurrentBib}

\bibitem [\protect \citeauthoryear {%
Jian%
}{%
Jian%
}{%
{\protect \APACyear {2017}}%
}]{%
jian2017recent}
\APACinsertmetastar {%
jian2017recent}%
\begin{APACrefauthors}%
Jian, Z.%
\end{APACrefauthors}%
\unskip\
\newblock
\APACrefYearMonthDay{2017}{}{}.
\newblock
{\BBOQ}\APACrefatitle {The recent trend of ethnic intermarriage in China: an
  analysis based on the census data} {The recent trend of ethnic intermarriage
  in china: an analysis based on the census data}.{\BBCQ}
\newblock
\APACjournalVolNumPages{The Journal of Chinese Sociology}{4}{1}{11}.
\PrintBackRefs{\CurrentBib}

\bibitem [\protect \citeauthoryear {%
Kozlowski%
, Taddy%
\BCBL {}\ \BBA {} Evans%
}{%
Kozlowski%
\ \protect \BOthers {.}}{%
{\protect \APACyear {2019}}%
}]{%
kozlowski2019geometry}
\APACinsertmetastar {%
kozlowski2019geometry}%
\begin{APACrefauthors}%
Kozlowski, A\BPBI C.%
, Taddy, M.%
\BCBL {}\ \BBA {} Evans, J\BPBI A.%
\end{APACrefauthors}%
\unskip\
\newblock
\APACrefYearMonthDay{2019}{}{}.
\newblock
{\BBOQ}\APACrefatitle {The geometry of culture: Analyzing the meanings of class
  through word embeddings} {The geometry of culture: Analyzing the meanings of
  class through word embeddings}.{\BBCQ}
\newblock
\APACjournalVolNumPages{American Sociological Review}{84}{5}{905--949}.
\PrintBackRefs{\CurrentBib}

\bibitem [\protect \citeauthoryear {%
Manzini%
, Lim%
, Tsvetkov%
\BCBL {}\ \BBA {} Black%
}{%
Manzini%
\ \protect \BOthers {.}}{%
{\protect \APACyear {2019}}%
}]{%
manzini2019black}
\APACinsertmetastar {%
manzini2019black}%
\begin{APACrefauthors}%
Manzini, T.%
, Lim, Y\BPBI C.%
, Tsvetkov, Y.%
\BCBL {}\ \BBA {} Black, A\BPBI W.%
\end{APACrefauthors}%
\unskip\
\newblock
\APACrefYearMonthDay{2019}{}{}.
\newblock
{\BBOQ}\APACrefatitle {Black is to Criminal as Caucasian is to Police:
  Detecting and Removing Multiclass Bias in Word Embeddings} {Black is to
  criminal as caucasian is to police: Detecting and removing multiclass bias in
  word embeddings}.{\BBCQ}
\newblock
\BIn{} \APACrefbtitle {{North American Chapter of the Association for
  Computational Linguistics: Human Language Technologies (NAACL-HLT)}.} {{North
  American Chapter of the Association for Computational Linguistics: Human
  Language Technologies (NAACL-HLT)}.}
\PrintBackRefs{\CurrentBib}

\bibitem [\protect \citeauthoryear {%
Michel%
\ \protect \BOthers {.}}{%
Michel%
\ \protect \BOthers {.}}{%
{\protect \APACyear {2011}}%
}]{%
michel2011quantitative}
\APACinsertmetastar {%
michel2011quantitative}%
\begin{APACrefauthors}%
Michel, J\BHBI B.%
, Shen, Y\BPBI K.%
, Aiden, A\BPBI P.%
, Veres, A.%
, Gray, M\BPBI K.%
, Team, G\BPBI B.%
\BDBL {}others%
\end{APACrefauthors}%
\unskip\
\newblock
\APACrefYearMonthDay{2011}{}{}.
\newblock
{\BBOQ}\APACrefatitle {Quantitative analysis of culture using millions of
  digitized books} {Quantitative analysis of culture using millions of
  digitized books}.{\BBCQ}
\newblock
\APACjournalVolNumPages{Science}{331}{6014}{176--182}.
\PrintBackRefs{\CurrentBib}

\bibitem [\protect \citeauthoryear {%
Mikolov%
}{%
Mikolov%
}{%
{\protect \APACyear {2013}}%
}]{%
mikolov2013efficient}
\APACinsertmetastar {%
mikolov2013efficient}%
\begin{APACrefauthors}%
Mikolov, T.%
\end{APACrefauthors}%
\unskip\
\newblock
\APACrefYearMonthDay{2013}{}{}.
\newblock
{\BBOQ}\APACrefatitle {Efficient estimation of word representations in vector
  space} {Efficient estimation of word representations in vector space}.{\BBCQ}
\newblock
\APACjournalVolNumPages{arXiv preprint arXiv:1301.3781}{3781}{}{}.
\PrintBackRefs{\CurrentBib}

\bibitem [\protect \citeauthoryear {%
Mikolov%
, Sutskever%
, Chen%
, Corrado%
\BCBL {}\ \BBA {} Dean%
}{%
Mikolov%
\ \protect \BOthers {.}}{%
{\protect \APACyear {2013}}%
}]{%
mikolov2013distributed}
\APACinsertmetastar {%
mikolov2013distributed}%
\begin{APACrefauthors}%
Mikolov, T.%
, Sutskever, I.%
, Chen, K.%
, Corrado, G\BPBI S.%
\BCBL {}\ \BBA {} Dean, J.%
\end{APACrefauthors}%
\unskip\
\newblock
\APACrefYearMonthDay{2013}{}{}.
\newblock
{\BBOQ}\APACrefatitle {Distributed representations of words and phrases and
  their compositionality} {Distributed representations of words and phrases and
  their compositionality}.{\BBCQ}
\newblock
\BIn{} \APACrefbtitle {{Proceedings of Advances in Neural Information
  Processing Systems (NeurIPS)}.} {{Proceedings of Advances in Neural
  Information Processing Systems (NeurIPS)}.}
\PrintBackRefs{\CurrentBib}

\bibitem [\protect \citeauthoryear {%
Papersnake%
}{%
Papersnake%
}{%
{\protect \APACyear {2023}}%
}]{%
ps2023pdnd}
\APACinsertmetastar {%
ps2023pdnd}%
\begin{APACrefauthors}%
Papersnake.%
\end{APACrefauthors}%
\unskip\
\newblock
\APACrefYearMonthDay{2023}{}{}.
\newblock
\APACrefbtitle {Peolple Daily News Dataset.} {Peolple daily news dataset.}
\newblock
\APAChowpublished
  {\url{https://huggingface.co/datasets/Papersnake/people_daily_news}}.
\PrintBackRefs{\CurrentBib}

\bibitem [\protect \citeauthoryear {%
Pennington%
, Socher%
\BCBL {}\ \BBA {} Manning%
}{%
Pennington%
\ \protect \BOthers {.}}{%
{\protect \APACyear {2014}}%
}]{%
pennington2014glove}
\APACinsertmetastar {%
pennington2014glove}%
\begin{APACrefauthors}%
Pennington, J.%
, Socher, R.%
\BCBL {}\ \BBA {} Manning, C\BPBI D.%
\end{APACrefauthors}%
\unskip\
\newblock
\APACrefYearMonthDay{2014}{}{}.
\newblock
{\BBOQ}\APACrefatitle {Glove: Global vectors for word representation} {Glove:
  Global vectors for word representation}.{\BBCQ}
\newblock
\BIn{} \APACrefbtitle {{Annual Conference on Empirical Methods in Natural
  Language Processing (EMNLP)}.} {{Annual Conference on Empirical Methods in
  Natural Language Processing (EMNLP)}.}
\PrintBackRefs{\CurrentBib}

\bibitem [\protect \citeauthoryear {%
Rosello%
}{%
Rosello%
}{%
{\protect \APACyear {1998}}%
}]{%
rosello1998declining}
\APACinsertmetastar {%
rosello1998declining}%
\begin{APACrefauthors}%
Rosello, M.%
\end{APACrefauthors}%
\unskip\
\newblock
\APACrefYear{1998}.
\newblock
\APACrefbtitle {Declining the Stereotype: Ethnicity and Representation in
  French Cultures} {Declining the stereotype: Ethnicity and representation in
  french cultures}.
\newblock
\APACaddressPublisher{}{Dartmouth College}.
\PrintBackRefs{\CurrentBib}

\bibitem [\protect \citeauthoryear {%
Walder%
}{%
Walder%
}{%
{\protect \APACyear {1989}}%
}]{%
walder1989social}
\APACinsertmetastar {%
walder1989social}%
\begin{APACrefauthors}%
Walder, A\BPBI G.%
\end{APACrefauthors}%
\unskip\
\newblock
\APACrefYearMonthDay{1989}{}{}.
\newblock
{\BBOQ}\APACrefatitle {Social change in post-revolution China} {Social change
  in post-revolution china}.{\BBCQ}
\newblock
\APACjournalVolNumPages{Annual Review of Sociology}{15}{1}{405--424}.
\PrintBackRefs{\CurrentBib}

\bibitem [\protect \citeauthoryear {%
Warriner%
, Kuperman%
\BCBL {}\ \BBA {} Brysbaert%
}{%
Warriner%
\ \protect \BOthers {.}}{%
{\protect \APACyear {2013}}%
}]{%
warriner2013norms}
\APACinsertmetastar {%
warriner2013norms}%
\begin{APACrefauthors}%
Warriner, A\BPBI B.%
, Kuperman, V.%
\BCBL {}\ \BBA {} Brysbaert, M.%
\end{APACrefauthors}%
\unskip\
\newblock
\APACrefYearMonthDay{2013}{}{}.
\newblock
{\BBOQ}\APACrefatitle {Norms of valence, arousal, and dominance for 13,915
  English lemmas} {Norms of valence, arousal, and dominance for 13,915 english
  lemmas}.{\BBCQ}
\newblock
\APACjournalVolNumPages{Behavior Research Methods}{45}{}{1191--1207}.
\PrintBackRefs{\CurrentBib}

\bibitem [\protect \citeauthoryear {%
Xie%
}{%
Xie%
}{%
{\protect \APACyear {2011}}%
}]{%
xie2011evidence}
\APACinsertmetastar {%
xie2011evidence}%
\begin{APACrefauthors}%
Xie, Y.%
\end{APACrefauthors}%
\unskip\
\newblock
\APACrefYearMonthDay{2011}{}{}.
\newblock
{\BBOQ}\APACrefatitle {Evidence-based research on China: A historical
  imperative} {Evidence-based research on china: A historical
  imperative}.{\BBCQ}
\newblock
\APACjournalVolNumPages{Chinese Sociological Review}{44}{1}{14--25}.
\PrintBackRefs{\CurrentBib}

\bibitem [\protect \citeauthoryear {%
Xu%
, Li%
\BCBL {}\ \BBA {} Chen%
}{%
Xu%
\ \protect \BOthers {.}}{%
{\protect \APACyear {2022}}%
}]{%
xu2022valence}
\APACinsertmetastar {%
xu2022valence}%
\begin{APACrefauthors}%
Xu, X.%
, Li, J.%
\BCBL {}\ \BBA {} Chen, H.%
\end{APACrefauthors}%
\unskip\
\newblock
\APACrefYearMonthDay{2022}{}{}.
\newblock
{\BBOQ}\APACrefatitle {Valence and arousal ratings for 11,310 simplified
  Chinese words} {Valence and arousal ratings for 11,310 simplified chinese
  words}.{\BBCQ}
\newblock
\APACjournalVolNumPages{Behavior Research Methods}{54}{1}{26--41}.
\PrintBackRefs{\CurrentBib}

\bibitem [\protect \citeauthoryear {%
Zeng%
\ \BBA {} Greenfield%
}{%
Zeng%
\ \BBA {} Greenfield%
}{%
{\protect \APACyear {2015}}%
}]{%
zeng2015cultural}
\APACinsertmetastar {%
zeng2015cultural}%
\begin{APACrefauthors}%
Zeng, R.%
\BCBT {}\ \BBA {} Greenfield, P\BPBI M.%
\end{APACrefauthors}%
\unskip\
\newblock
\APACrefYearMonthDay{2015}{}{}.
\newblock
{\BBOQ}\APACrefatitle {Cultural evolution over the last 40 years in China:
  Using the Google Ngram Viewer to study implications of social and political
  change for cultural values} {Cultural evolution over the last 40 years in
  china: Using the google ngram viewer to study implications of social and
  political change for cultural values}.{\BBCQ}
\newblock
\APACjournalVolNumPages{International Journal of Psychology}{50}{1}{47--55}.
\PrintBackRefs{\CurrentBib}

\end{thebibliography}

\end{CJK*}
\end{document}